\documentclass[sigconf]{acmart}

\usepackage{listings}
\usepackage{tabularx}
\usepackage{graphicx}
\usepackage{epsfig}
\usepackage{amssymb}
\usepackage{amsmath}
\usepackage{hyperref}
\usepackage{caption}
\usepackage{subcaption}
\usepackage{epstopdf}
\usepackage{float}
\usepackage{graphicx}
\usepackage{multirow}
\usepackage{amsfonts}
\usepackage{booktabs}

\usepackage{xstring}
\usepackage{array}
\usepackage{ragged2e}
\newcolumntype{P}[1]{>{\RaggedRight\hspace{0pt}}p{#1}}

\copyrightyear{2019} 
\acmYear{2019} 
\setcopyright{acmlicensed}
\acmConference[e-Energy '19]{Proceedings of the Tenth ACM International Conference on Future Energy Systems}{June 25--28, 2019}{Phoenix, AZ, USA}
\acmBooktitle{Proceedings of the Tenth ACM International Conference on Future Energy Systems (e-Energy '19), June 25--28, 2019, Phoenix, AZ, USA}
\acmPrice{15.00}
\acmDOI{10.1145/3307772.3328284}
\acmISBN{978-1-4503-6671-7/19/06}

\def\BibTeX{{\rm B\kern-.05em{\sc i\kern-.025em b}\kern-.08emT\kern-.1667em\lower.7ex\hbox{E}\kern-.125emX}}
    
\begin{document}

\title{Energy Predictive Models with Limited Data\\ using Transfer Learning}

\author{Ali Hooshmand}
\email{ahooshmand@nec-labs.com}
\affiliation{%
\institution{NEC Laboratories America, CA}
}

\author{Ratnesh Sharma}
\email{ratnesh@nec-labs.com}
\affiliation{%
	\institution{NEC Laboratories America, CA}
}

%
\begin{abstract}
In this paper, we consider the problem of developing predictive models with limited data for energy assets such as electricity loads, PV power generations, etc. We specifically investigate the cases where the amount of historical data is not sufficient to effectively train the prediction model. We first develop an energy predictive model based on convolutional neural network (CNN) which is well suited to capture the interaday, daily, and weekly cyclostationary patterns, trends and seasonalities in energy assets time series. A transfer learning strategy is then proposed to address the challenge of limited training data. We demonstrate our approach on a usecase of daily electricity demand forecasting. 
we show practicing the transfer learning strategy on the CNN model results in significant improvement to existing forecasting methods. 
\end{abstract}

%
%
%


\acmYear{2019}\copyrightyear{2019}
\setcopyright{acmlicensed}
\acmConference[e-Energy '19]{e-Energy '19: Proceedings of the Tenth ACM International Conference on Future Energy Systems}{June 25--28, 2019}{Phoenix, AZ, USA}
\acmBooktitle{e-Energy '19: Proceedings of the Tenth ACM International Conference on Future Energy Systems, June 25--28, 2019, Phoenix, AZ, USA}
\acmPrice{15.00}
\acmDOI{10.1145/3307772.3328284}
\acmISBN{978-1-4503-6671-7/19/06}
%

%
\maketitle
\section{Introduction}
Various prediction models are utilized in power and energy systems such as electricity load forecasting, PV prediction, wind generation estimation, 
etc. 
%
Applying machine learning techniques to develop prediction models is a well-established and active research field attracting considerable researchers motivated by the new advances in computational technologies and machine learning techniques.
Current algorithms require to have enough historical data of the energy assets and customers for training models. 
In \cite{yona2008application}, a recurrent neural network is designed to predict 24 hour ahead PV power generation. Provided by Japan Meteorological Business Support Center, two years of meteorogical data was utilized in their research to train the deep learning model. 
Authors in \cite{wu2010hybrid} proposed a hybrid day-ahead energy price forecasting model based on time-series and adaptive wavelet neural networks. Their model is applied to forecast the electricity price for PJM electricty market and used one year of historical recorded PJM data to be trained. In \cite{mabel2008analysis}, a deep learning model was proposed for wind power generation estimation where three years of field data of seven wind farms in India were used to develop the model. Hence, similar to other use-cases of modern machine learning techniques, having access to large amount of data is crucial in training accurate models. 
There are, however, many applications in energy systems for which large historical data might not be available. For example, new customers might have been added or new facilities or assets might have been installed. In such cases, lack of sufficient data results in underperforming machine learning models. This constitutes the main challenge of this research which is establishing predictive platforms in settings where limited data is available for customers. We propose to compensate for small historical data by transferring features from models that are pre-trained with distinct data of similar nature. In doing so, we first develop a a deep learning model based on convolutional neural networks (CNN) for energy time series data with daily seasonality. It is able to consider the trends and seasonality of energy data in its prediction algorithm. We chose the CNN model over Recurrent Neural Networks (RNN), since it can efficiently extract correlations in electricity power data at different time-scales. Several other papers have also reported case-studies where CNNs outperform RNNs for time-series data~\cite{lee2009unsupervised,kalchbrenner2014convolutional,bai2018empirical}. We then propose a transfer learning procedure for pre-training the model on historical data and fine-tuning it with limited data of new customers. Our experimental results show that the proposed method outperforms current techniques for load forecasting. Authors in ~\cite{ribeiro2018transfer} has also applied transfer learning to forecast building energy consumption. They leverage the consumption data of similar buildings, and augment it to limited data of target building for which the model is trained. The augmented data is then passed through the steps of Hephaestus (their proposed transfer learning method), namely Time Series Adaptation and Non-Temporal Domain Adaptation. Their strategy is implemented on the input data and then the processed data is used to train a machine learning model. However, in our proposed method the transfer learning strategy is implemented on the way we train the CNN model.  

The rest of this paper is organized as follows. Section~\ref{sec:CnnModel} presents our design choice for time series prediction model based on convolutional neural nets. In section~\ref{sec:TransferLearning}, we propose our transfer learning approach using the CNN model. In section~\ref{sec:CaseStudy}, we implement our designed transfer learning technique on load forecasting problem as a case study and provide its experimental results, and Section~\ref{sec:Conclusion} concludes the paper.

\section{Proposed CNN-Based Time Series predictive model}\label{sec:CnnModel}

In this section, we describe the prediction model goal, our approach of preprocessing energy time series data as input for predictive model, and provide the structure of CNN-based predictive model.

\subsection{Goal}
The objective in designing the prediction model is to use the historical recorded energy data to forecast the next 24 hours energy profile as the model output. The model could be applied for energy time-series predictions such as electrical load, PV generation, and wind power production. The prediction should be accomplished so as
to minimize prediction error at each time instance. We choose
the mean absolute error (MAE) as the training criterion:
$MAE = (1/N) \sum_{i=1}^{N}|x_i - \hat{x}_i|$ \\
where $x_i$ is the actual value, $\hat{x}_i$ is the predicted value, and $N$ is number of samples.  

\subsection{Energy Time Series Data}

\begin{figure}[H]
	\centering
	\includegraphics[width=0.80\linewidth]{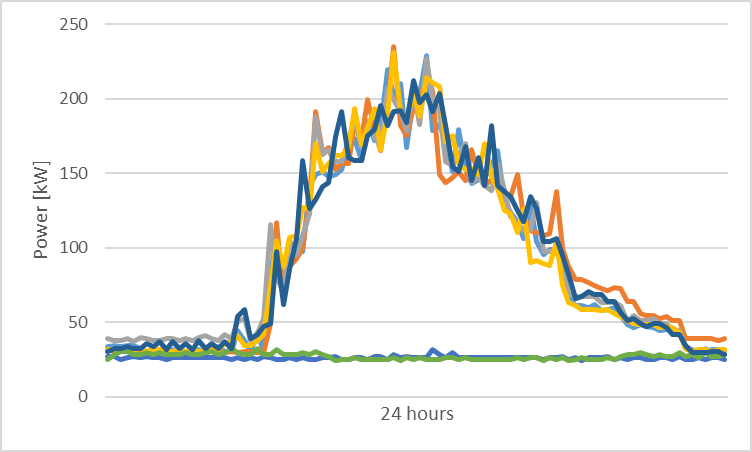}
	\vspace{-10pt}
	\caption{A sample of load consumption for 7 days of a week}
	\label{fig:dailyLoad}
\end{figure}

Energy data, e.g., electricity, PV, wind, etc., are usually in the form of time-series with daily cyclic patterns. 
It means for most of the applications, data profiles such as electricity load, PV generation, and wind turbine generation repeat their pattern every 24 hours. An example of daily load consumption of a school facility for 7 days of a week is shown in Figure~\ref{fig:dailyLoad}. The two flat load profiles represent the electricity usage of weekend days.   


\subsection{Model Inputs}


The main input of the model is past daily data profiles. We use data of past four weeks to forecast energy profile of the next day. It is assumed the past four weeks data provides enough information about interaday trends, and cyclo-stationary seasonality patterns with daily and weekly periods. For each day, we reshape data of past four weeks into a three dimensional matrix of size $24$ (it should be stated before that data is sampled once every 60 min) by $7$ (number of days in a week) by $4$ (number of weeks), to align data of consecutive hours, days and weeks next to each other.
Note that the number of samples per day might vary based on the sampling time of collected data and the requirements of application the prediction profile is used for. 



\begin{table}[]
	\centering
	\caption{Structure of convolutional network.}
	\vspace{-10pt}
	\label{tab:conv}
	\begin{tabular}{|l|c|} 
		\hline
		{\bf Layer} & {\bf Output Shape} \\
		\hline
		\hline
		Input & $(96, 7, 4)$ \\
		\hline
		Convolutional with $32$ filters of size $(3,3)$ & $(96, 7, 32)$ \\
		Relu, Batch Norm, Maxpool (4,1) & $(24, 7, 32)$ \\
		
		\hline
		Convolutional with $64$ filters of size $(3,3)$ & $(24, 7, 64)$ \\
		Relu, Batch Norm, Maxpool (2,1) & $(12, 7, 64)$ \\
		
		\hline
		Convolutional with $128$ filters of size $(3,3)$ & $(12, 7, 128)$ \\
		Relu, Batch Norm, Maxpool (2,1) & $(6, 7, 128)$ \\
		
		\hline
		Convolutional with $256$ filters of size $(3,3)$ & $(6, 7, 256)$ \\
		Relu, Batch Norm, Maxpool (2,2) & $(3, 3, 256)$ \\
		
		\hline
		Convolutional with $512$ filters of size $(3,3)$ & $(3, 3, 512)$ \\
		Relu, Batch Norm, Maxpool (2,2) & $(1, 1, 512)$ \\
		
		\hline
		Dense, Relu, Dropout(0.5) & $(1, 512)$ \\
		
		\hline
		Linear(output) & $(1,96)$\\
		
		\hline

	\end{tabular}
\end{table}

\subsection{Model Structure}
The three dimensional matrix of historical data is formed in such a way that the hourly, daily and weekly correlations are aligned across different axises and, thus, a convolutional neural network is a suitable tool to extract the spatial correlations of such a  data. The designed CNN model is inspired by VGG network  which was developed for classifying ImageNet datatset~\cite{simonyan2014very} (the winner of Image Net Large Scale Visual Recognition Challange, ILSVRC in 2014~\cite{ILSVRC}) and composed of five convolutional, one dense layer with relu activation and one linear output layer. The model output is the day-ahead energy time series profile. Hence, the model loss function is defined as follows:\\
$Loss = \sum_{i=1}^n \sum_{t=0}^T (P_i(t) - \hat{P}_i(t))^2$ \\
Table~\ref{tab:conv} provides the details of designed CNN model.


\begin{figure}
	\centering
	\includegraphics[width=0.9\linewidth]{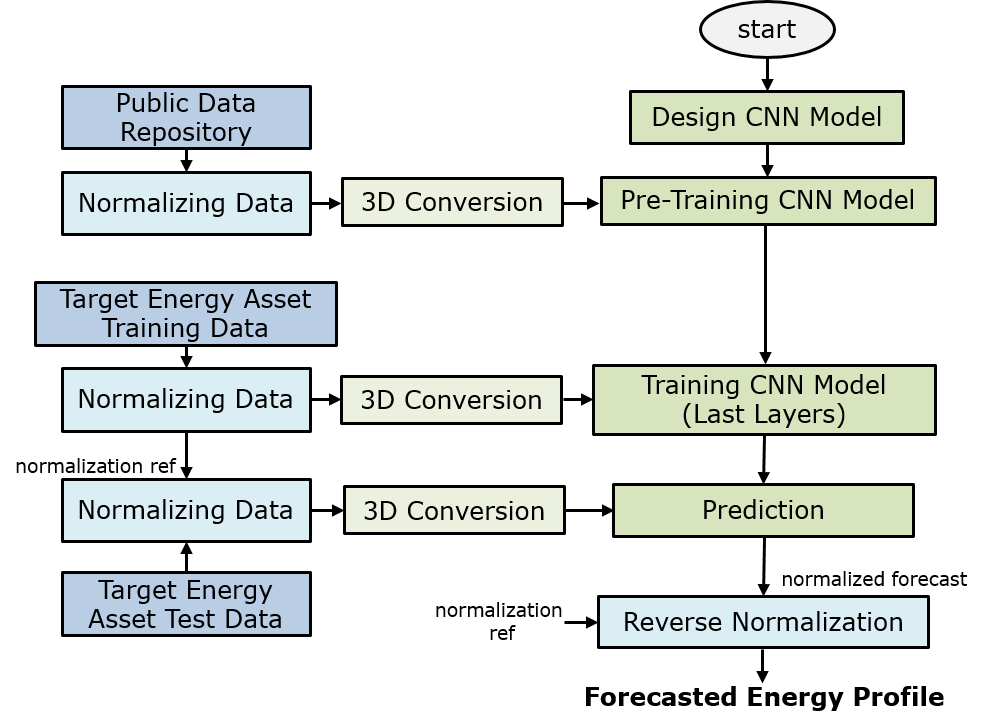}
	\vspace{-10pt}
	\caption{Proposed predictive model development flowchart for energy assets with limited historical data. 
	}
	\label{fig:flowchart}
\end{figure}

\section{Transfer Learning Method}\label{sec:TransferLearning}
With sufficient historical energy data, the CNN model could be effectively trained to yield desirable accuracy. However, in many cases, we do  not have access to large amount of data. 
To address this problem, we propose a transfer learning procedure through the steps which are described in this section. Transfer learning is a general concept in machine learning in which the knowledge learned from a dataset or a problem is transferred to a different but related problem for which it is expensive or impossible to collect enough training data~\cite{pan2010survey}. 
%
\vspace{-8pt}
\subsection*{Step 1: Public Datasets}
At the first step, we collect publicly available energy data for different types of customer categories and energy assets. This data could be obtained from open source datasets. As an example, Open PV Project~\cite{openpv} is a collaboration between Department of Energy (DOE), industry, and public community to create a comprehensive database for storing available PV generation data in the United States. This project is established by NREL (National Renewable Energy Laboratory)~\cite{nrel}. 
Using the public datasets, we can train the CNN model for the type of energy asset under study.
%
\vspace{-8pt}
\subsection*{Step 2: Normalization}
One problem with using public datasets is that the scale of data could be significantly different from the target application. 
In order to correct the scaling differences, we normalize the public dataset so as to enable the model to efficiently learn the trends and seasonality of time-series profiles which are useful for development of target customer predictive model. 
\vspace{-8pt}
\subsection*{Step 3: Pre-Training}
The designed CNN model is trained using the pre-processed data of public repository datasets. The pre-training step helps us to learn from public data the specific features that are common across energy assets with the same type as the target asset. The gained knowledge is transfered to our target prediction model in the next step.  
\vspace{-8pt}
\subsection*{Step 4: Training}
The trained model in \textit{Step 3} does not yet result in the best expected performance for the target energy asset since it has not been trained with its specific data and scale. To improve the accuracy, we fine-tune the model on the available limited data of the target user. however, in order to preserve the features that are learned using the large public datasets, we freeze the convolutional layers and only retrain the last fully-connected layers. This approach customizes the model to new data while keeping the original features. Notably, since the number of parameters in last two layers is significantly smaller than the convolutional layers, the small data of target asset suffices to achieve a desirable performance. To this end, the limited customer data is randomly split into training and test sets. The train set is normalized and the normalization reference (the maximum value of training data) is recorded for the prediction step. 
After fine-tuning, the model is evaluated on test data of the target asset.   
\vspace{-8pt}
\subsection*{Step 5: Prediction}
To perform forecasting, the test dataset is first normalized using the normalization reference recorded in training step. Then, it is passed to the trained CNN model for prediction. The model output is then reversly normalized (using the normalization reference) at the final stage which results in the forecasted profile for the next 24-hour generation or consumption of target energy asset.

By pretraining the CNN model based on public related-domain dataset, we could train the weights of convolutional layers as well as initialize the weights of last fully connected layers. It lets us to use the limited data of target customer to fine tune the weights of last layer and achieve better performance. The proposed strategy is summarized in the flowchart shown in Figure~\ref{fig:flowchart}.

\section{Case Study: \\Electricity Load Forecasting}\label{sec:CaseStudy}
\begin{figure}[t]
	\centering
	\includegraphics[width=1\linewidth]{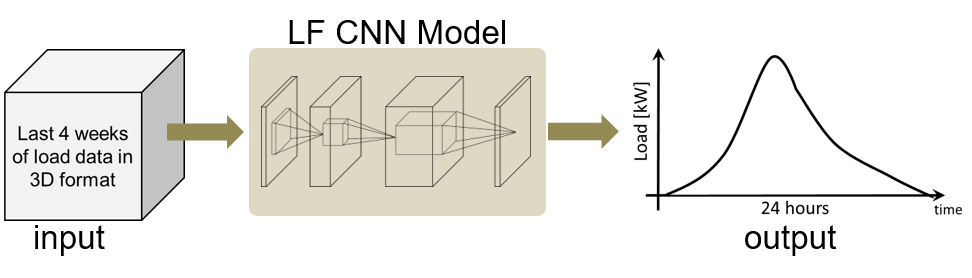}
	\vspace{-25pt}
	\caption{Input and output for load forecast CNN model.}
		\vspace{-10pt}
	\label{fig:input_output}
\end{figure}

Day-ahead load forecasting in power and energy systems refers to the prediction of future electrical demand for the next 24 hours ~\cite{gross1987short}. 
The load forecasting (LF) model input is chosen to be the load data for the past four weeks and the model output is forecasted profile for the next 24 hours of load consumption as shown in Figure~\ref{fig:input_output}. 

We aim to develop the LF model for customers who have limited historical recorded data insufficient for training the CNN model. 
To prepare the model, we will follow the steps presented in section~\ref{sec:TransferLearning}. We start by utilizing the publicly available electrical load data for different types of customers. Here, this data is obtained from the open source dataset in UCI repository~\cite{uci}. The dataset contains the four years electricity consumption of about 370 clients. 
\begin{figure}[h]
	\centering
		\centering
		\includegraphics[width=0.8\linewidth]{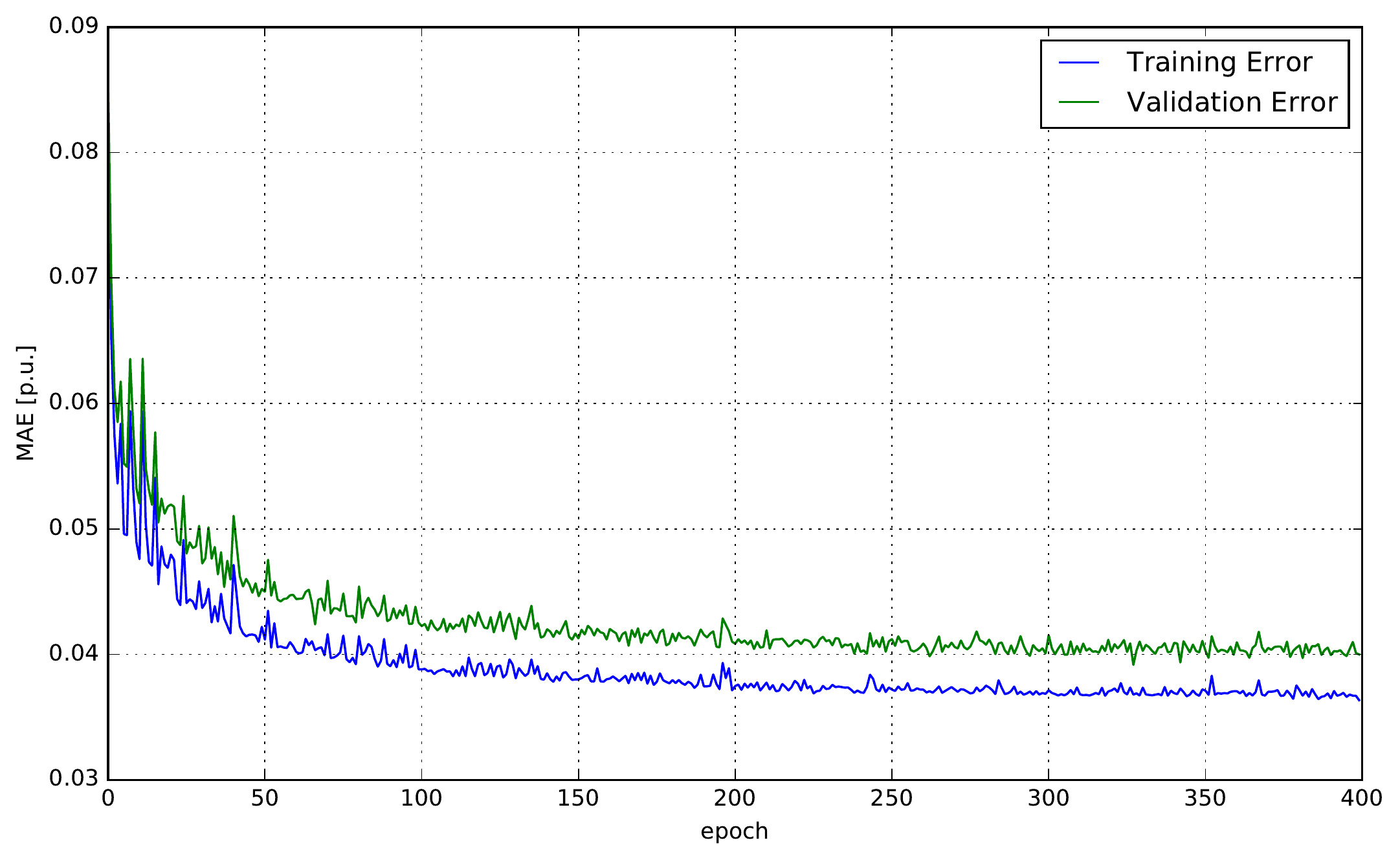}
		\vspace{-12pt}
	\caption{Training and validation mean absolute errors (MAE) for load prediction during 400 epochs of training.}
		\vspace{-10pt}
	\label{fig:train_valid_error}
\end{figure}
\begin{figure*}[t]
	\vspace{-8pt}
	\centering
	\includegraphics[width=.9\linewidth]{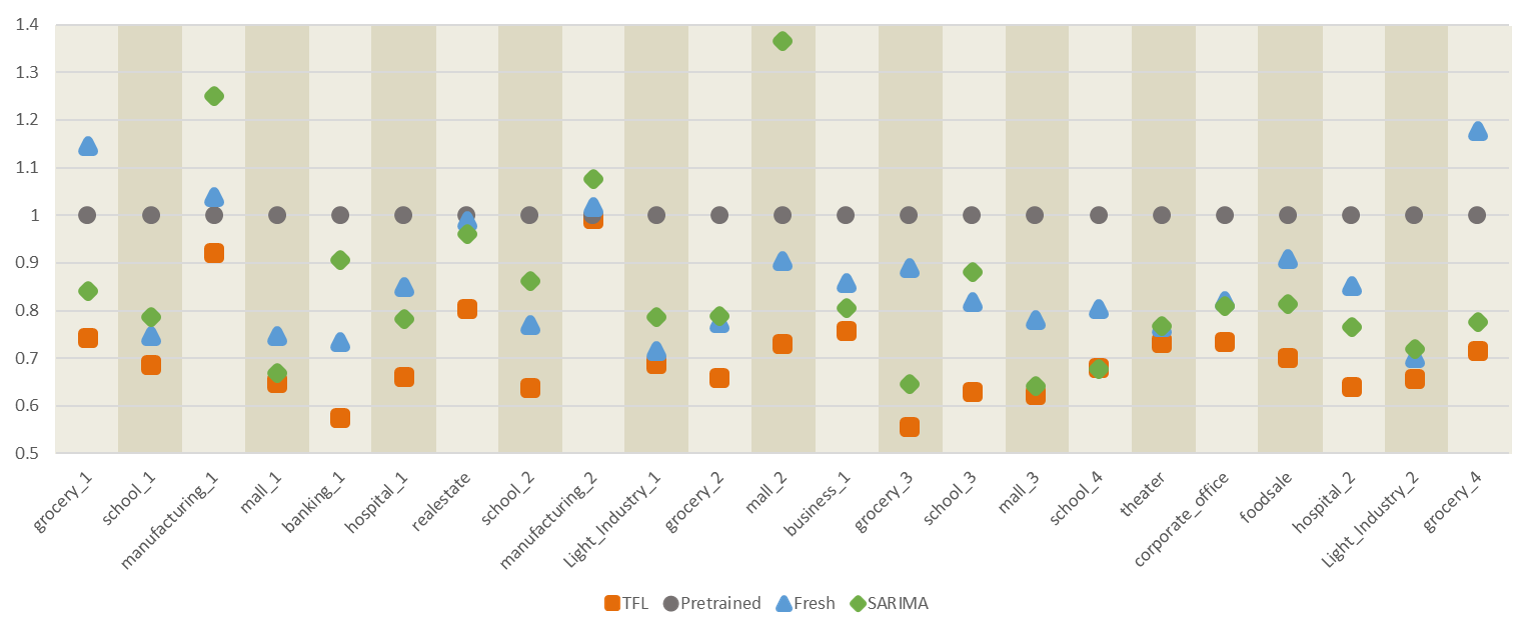}
	\vspace{-15pt}
	\caption{Normalized averaged MAE (N.A.MAE) comparison for proposed transfer learning strategy (TFL), Pre-trained, Fresh, and SARIMA models for 23 C\&I customers in different segments (normalized based on Pre-trained model averaged MAE value).}
	\label{fig:comparison}
\end{figure*} 

To eliminate the local features, each customer load in dataset is normalized based on it maximum value that projects the scale of all loads into the range of 0 to 1. The dataset is randomly split into two sets of training data (70\%) and validation data (30\%). 
The model is trained (pre-training step described in section~\ref{sec:TransferLearning}) to minimize the sum of MAE of load forecast error. Figure~\ref{fig:train_valid_error} shows the training and validation errors for the load forecast during 400 epochs of training. 
The model with the best validation error will be saved for the training step explained in section~\ref{sec:TransferLearning}. By pre-training step, the model learned the electrical load common features from normalized version of UCI public data set. 

The model retraining is performed using the available load data of the target customer for which the forecasting model is prepared.   
The selected customers have recorded data for about five months from which four months worth of data is used for retraining the saved model in previous step. 
The rest of data is kept as the test set for assessing the prediction accuracy of the final model. 

To evaluate the performance of proposed transfer learning strategy on designed CNN model, we compare our proposed method with the following medels:
\begin{itemize}
	\item SARIMA 
	model as a classic time-series prediction technique ~\cite{brockwell2002introduction} which does not require a lot data to be trained due to its linear and simple structure. 
	\item Fresh CNN model in which the designed CNN model in section~\ref{sec:CnnModel} is trained only using the target load data without using the public dataset and pre-training step.
	\item Pre-Trained model in which the best model saved during pre-training step in section~\ref{sec:TransferLearning} is directly used for load forecasting of target customer without using its limited historical data.
\end{itemize}

\bgroup
\def\arraystretch{1.0}
\begin{table}
	\centering
	\caption{Percentage improvement of proposed transfer learning approach with respect to each model. 
	}
\vspace{-10pt}
	\label{tab:comp}
	\begin{tabular}{|P{2.4cm}|P{1.7cm}|P{1.6cm}|P{1.6cm}|} 
		\hline
		Total Normalized Averaged MAE (TNA.MAE) & {\bf {\scriptsize Pre-trained Model (TNA.MAE: 1)}} & {\bf {\scriptsize Fresh Model (TNA.MAE: 0.84)}} & {\bf {\scriptsize SARIMA Model (TNA.MAE: 0.87)}} \\
			\hline
			\hline
			{\bf {\scriptsize Transfer Learning Method (TNA.MAE:~0.7)}} & 30\% & 17\% & 20\% \\
			\hline
		\end{tabular}
		\vspace{-16pt}
	\end{table}

\egroup

The proposed transfer learning strategy was implemented for 23 randomly selected commercial and industrail (C\&I) customers with limited data in different segments such as grocery, school, manufacturing, mall, banking, hospital, theater, etc. A sample of daily load data of these customers can be seen in Figure ~\ref{fig:sampleLoad} of Appendix section. Please note the difference in load scale of different customers. For each customer, the averaged MAE of 
test days is calculated for each method separately. 
Finally daily MAEs are averaged over all 
test days. Since the load scale of selected customers varies, the MAE results are normalized based on the averaged MAE value of Pre-trianed model to better observe and compare the performance of different models. We call this value NA.MAE (normalized averaged MAE). Figure~\ref{fig:comparison} illustrates the performance of the developed method together with the performance of SARIMA, Fresh model, and Pre-trained model. As it can be seen, the proposed transfer learning strategy outperforms other models for all C\&I loads as it always has the lowest NA.MAE value for all examined C\&I loads. The absolute value of errors can also be found in Table ~\ref{tab:mae} in Appendix section. 

Note that the performance of Fresh model is not satisfactory. However, the results are consistently improved when the limited training data is used to fine-tune the model that is pretrained with public datasets (TFL orange squares in Figure~\ref{fig:comparison}). 

We also calculate the total normalized averaged MAE (TNA.MAE) for each model which is the average of NA.MAE values over all 23 load categories. Table~\ref{tab:comp} provides the summary of TNA.MAE results. For example, the TNA.MAE of SARIMA model is 0.87, meaning that the total error of SARIMA model for all 
test days and all 23 customers is 87\% of Pre-trained model's error. This value for our proposed method is 0.70. The second row of the table also shows the percentage improvement of our transfer learning approach with respect to each model. It performed 30\%, 17\%, and 20\% ($(0.87-0.70)/0.87$) better than Pre-trained, Fresh, and SARIMA models respectively. 
Better performance of our proposed model in comparison with the Fresh model (17\%) proves common features learned and transferred from public dataset has helped our approach to improve prediction results. Also, better prediction outcome of our approach compared to Pre-trained model (30\%) demonstrates the necessity of retraining the last dense layers to learn the local features and behavior of target customer limited load data.

\section{Conclusion}\label{sec:Conclusion}

In this paper, we proposed a procedure for designing energy predictive models in small-data regimes. To this end, we designed a convolutional network for time-series forecasting and used ideas from transfer learning to improve the accuracy. The proposed framework was implemented on the usecase of short-term electricity load forecasting in which the model predicts the next 24 hours of electricity demand based on the energy consumption of past four weeks. The simulation studies were conducted for 23 C\&I customers and showed our approach consistently results in lower error compared to SARIMA, Fresh CNN model, and Pre-trained CNN model. 
The proposed approach could help energy customers with limited data to develop well-trained predictive tools.

%
\bibliographystyle{ACM-Reference-Format}
\bibliography{bibfile}

\section*{Appendix}

\begin{figure}[H]
	\centering
	\centering
	\includegraphics[width=0.8\linewidth]{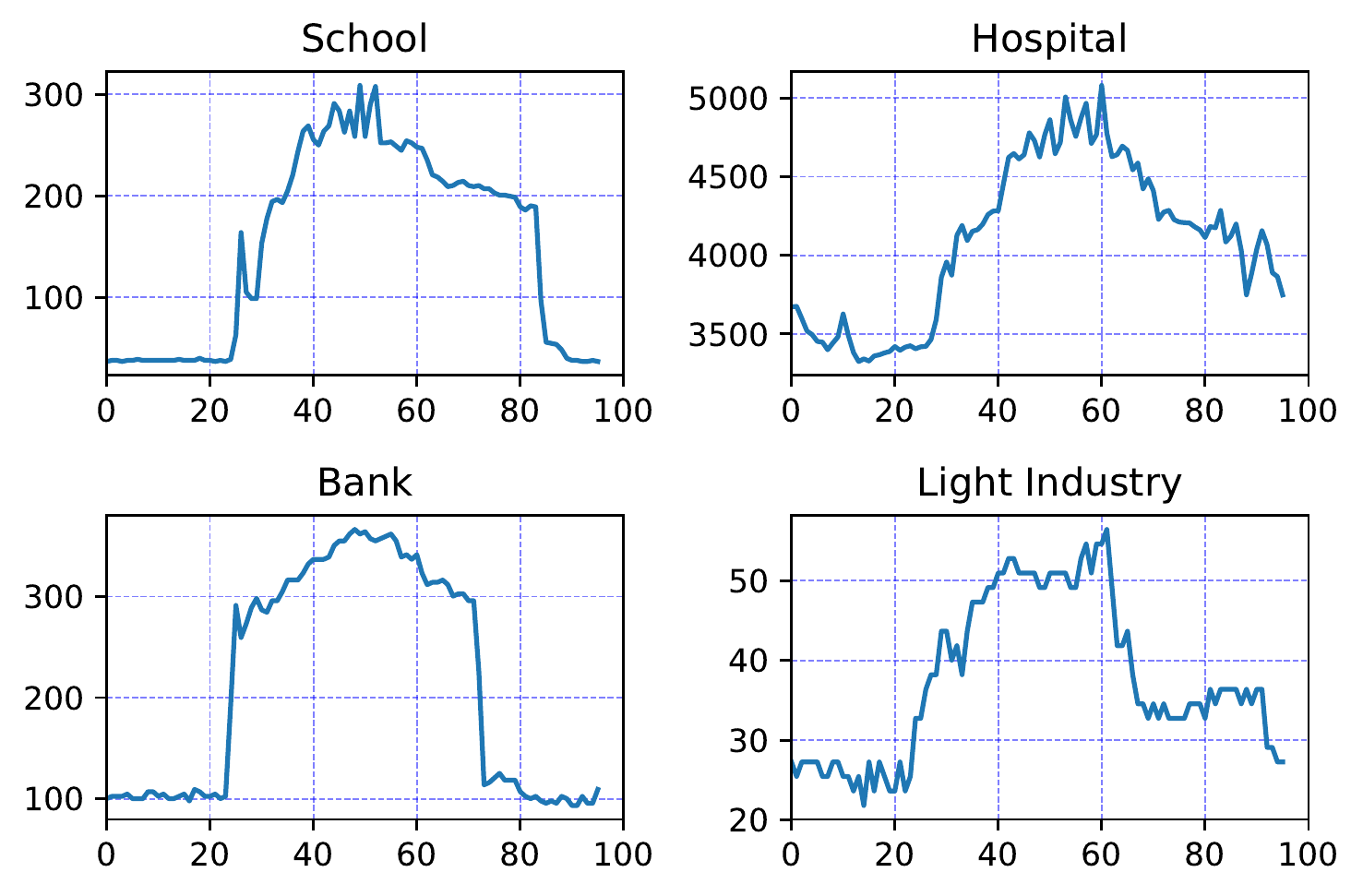}
	\caption{Daily load samples of C\&I customers. The load unit is in kW and data is sampled every 15 minutes (96 samples per day).}
	\label{fig:sampleLoad}
\end{figure}

Figure~\ref{fig:sampleLoad} illustrates daily load samples of four randomly selected customers from school, hospital, light industry, and banking segments. The unit is in kW and load is sampled every 15 minutes (96 samples per day).
\vspace{-40pt}

\vspace{15pt}
Table~\ref{tab:mae} shows the day-ahead load forecasting mean absolute error, averaged over all test days, for selected 23 C\&I customers. 
\begin{table}[H]
	\caption{Averaged MAE [kW] of test days for 23 C\&I customers for evaluated predictive models}  
	\label{tab:mae}
	\small 
	\centering 
	\begin{tabular}{lcccr} 
		\toprule[\heavyrulewidth]\toprule[\heavyrulewidth]
		\textbf{} & \textbf{} & \textbf{Predictive Models} & \textbf{}& \textbf{} \\ 
\midrule
		\textbf{Load Name} & \textbf{TFL} & \textbf{Pre-trained} & \textbf{Fresh}& \textbf{SARIMA} \\ 
		\midrule
		\multirow{1}{*}{grocery\_1} & 6.1 & 8.3 & 9.5 & 7.0 \\
		\midrule
		\multirow{1}{*}{school\_1} & 17.9 & 26.2 & 19.6 & 20.6\\
		\midrule
		\multirow{1}{*}{manufacturing\_1} & 37.5 & 40.8 & 42.4 & 51.0 \\
		\midrule
		\multirow{1}{*}{mall\_1} & 31.3 & 48.4 & 36.1 & 34.4 \\
		\midrule
		\multirow{1}{*}{banking\_1} & 32.4 & 56.4 & 41.4 & 51.1 \\
		\midrule
		\multirow{1}{*}{hospital\_1} & 161.8 & 244.9 & 207.9 & 191.5 \\
		\midrule
		\multirow{1}{*}{realestate} & 38.1 & 47.4 & 46.9 & 45.6 \\
		\midrule
		\multirow{1}{*}{school\_2} & 41.6 & 63.4 & 50.3 & 56.4 \\
		\midrule
		\multirow{1}{*}{manufacturing\_2} & 232.3 & 234.4 & 238.5 & 252.1 \\
		\midrule
		\multirow{1}{*}{light industry\_1} & 4.4 & 6.4 & 4.6 & 5.1 \\
		\midrule
		\multirow{1}{*}{grocery\_2} & 8.6 & 13.1 & 10.1 & 10.3 \\
		\midrule
		\multirow{1}{*}{mall\_2} & 279.3 & 383.1 & 346.0 & 522.8 \\
		\midrule
		\multirow{1}{*}{business\_1} & 36.0 & 47.6 & 40.8 & 38.3 \\
		\midrule
		\multirow{1}{*}{grocery\_3} & 5.5 & 9.9 & 8.8 & 6.4 \\
		\midrule
		\multirow{1}{*}{school\_3} & 21.4 & 34.0 & 27.8 & 29.9 \\
		\midrule
		\multirow{1}{*}{mall\_3} & 29.5 & 47.4 & 37 & 30.5 \\
		\midrule
		\multirow{1}{*}{school\_4} & 28.1 & 41.3 & 33.1 & 28.0 \\
		\midrule
		\multirow{1}{*}{theater} & 17.4 & 23.9 & 18.3 & 18.3 \\
		\midrule
		\multirow{1}{*}{corpoffice\_office} & 58.0 & 78.9 & 64.7 & 63.9 \\
		\midrule
		\multirow{1}{*}{foodsales} & 6.7 & 9.5 & 8.7 & 7.8 \\
		\midrule
		\multirow{1}{*}{hospital\_2} & 165.0 & 258.1 & 219.6 & 197.4 \\
		\midrule
		\multirow{1}{*}{light industry\_2} & 22.4 & 34.1 & 23.9 & 24.6 \\
		\midrule
		\multirow{1}{*}{grocery\_4} & 7.1 & 10.0 & 11.8 & 7.8 \\
		
		\bottomrule[\heavyrulewidth] 
	\end{tabular}
\end{table}

\end{document}